# Empirical Comparison of Lightweight Forecasting Models for Seasonal and Non-Seasonal Time Series

Thanh Son Nguyen

*IT Department, HCMC University of Technology and Education, 01 Vo Van Ngan, Thu Duc City
Ho Chi Minh City, 70000, Vietnam*[1]
*sonnt@hcmute.edu.vn*

Dang Minh Duc Nguyen

*IT Department, International University-VNU-HCM, Quarter 6, Linh Trung Ward, Thu Duc City
Ho Chi Minh City, 70000, Vietnam*
*itcsiu21050@student.hcmiu.edu.vn*

Van Thanh Nguyen

*Indigo Technology Systems JSC, Lot T2-4, Saigon Hi-Tech Park, Thu Duc City
Ho Chi Minh City, 70000, Vietnam*
*van-thanh.nguyen@indigo-jsc.com*

**Abstract**. Accurate time series forecasting is essential in many real-time applications that demand both high predictive accuracy and computational efficiency. This study provides an empirical comparison between a Polynomial Classifier (PC) and a Radial Basis Function Neural Network (RBFNN) across four real-world time series datasets (weather conditions, gold prices, crude oil prices, and beer production volumes) that cover both seasonal and non-seasonal patterns. Model performance is evaluated by forecasting accuracy (using Mean Absolute Error (MAE), Root Mean Squared Error (RMSE), and Coefficient of Variation of RMSE (CV(RMSE))) and computational time to assess each model's viability for real-time forecasting. The results show that the PC yields more accurate and faster forecasts for non-seasonal series, whereas the RBFNN performs better on series with pronounced seasonal patterns. From an interpretability standpoint, the polynomial model offers a simpler, more transparent structure (in contrast to the neural network's black-box nature), which is advantageous for understanding and trust in real-time decision-making. The performance differences between PC and RBFNN are statistically significant, as confirmed by paired t-tests and Wilcoxon signed-rank tests. These findings provide practical guidance for model selection in time series forecasting, indicating that PC may be preferable for quick, interpretable forecasts in non-seasonal contexts, whereas RBFNN is superior for capturing complex seasonal behaviors.

**Keywords**: Time series forecasting, polynomial classifier, Radial Basis Function Neural Network, Comparative Analysis.

## 1. Introduction

Time series forecasting plays a critical role in various domains such as finance, energy, economics, and climate science. Accurate forecasting enables better planning, resource allocation, and risk management. As such, developing and evaluating forecasting models remains an important research direction.

In recent years, there has been a growing interest in lightweight forecasting models that offer a favorable balance between prediction accuracy, interpretability, and computational efficiency. These models are particularly attractive in scenarios such as embedded systems, edge computing, or real-time applications, where deep learning models may be too resource-intensive. Polynomial classifiers and shallow neural networks like RBFNN represent two popular lightweight alternatives that can be deployed with limited computational overhead.

Over the past decades, a wide range of models have been proposed, from classical statistical methods to machine learning and deep learning approaches. Among them, neural network-based models have shown significant success in capturing nonlinear dependencies in time series data. For instance, Lin et al. applied multilayer perceptrons (MLPs) and radial basis function neural networks (RBFNNs) to forecast electricity price fluctuations in the PJM market.[1] Similarly, fuzzy control combined with RBFNN has been proposed for short-term load forecasting in power systems.[2] Catalao et al. developed MLP-based models for electricity price prediction in California and Spain,[3] while Anbazhagan et al. employed recurrent neural networks to forecast electricity prices, outperforming traditional MLPs.[4] More recently, deep learning models such as long short-term memory (LSTM) and convolutional neural networks (CNNs) have been utilized for day-ahead electricity price forecasting and financial data prediction.[5,6]

In parallel, polynomial classifiers—regarded as simplified higher-order neural networks—have demonstrated strong performance in classification, recognition, and regression tasks due to their compact architecture and low computational complexity.[7] Despite these advantages, their application in time series forecasting remains limited. The first-order polynomial classifier, in particular, has shown promising results in stock price prediction,[8] suggesting potential utility in broader forecasting contexts.

This paper presents an experimental comparative study between a first-order polynomial classifier and RBFNN for time series forecasting. RBFNN is selected due to its strong function approximation capabilities and relatively fast learning rate, making it a reliable benchmark model. We assess both models using real-world datasets from diverse domains and evaluate them based on predictive accuracy and execution time.

The results show that while performance varies across datasets, the polynomial classifier achieves comparable or superior accuracy on non-seasonal data and offers slightly faster execution time, highlighting its potential as a lightweight and interpretable forecasting tool.

The remainder of this paper is structured as follows. Section 2 reviews RBFNN. Section 3 provides background on polynomial classifiers. Section 4 describes the experimental setup and datasets. Section 5 presents and discusses the results. Section 6 concludes the paper.

## 2. RBFNN model

The Radial Basis Function Neural Network (RBFNN) is a class of artificial neural networks (ANNs) widely used for function approximation, classification, and time series forecasting. It is particularly known for its universal approximation capability and faster learning speed compared to traditional multilayer perceptron (MLP) models.

A standard RBFNN consists of three layers: an input layer, a hidden layer, and an output layer. The hidden layer employs radial basis functions as activation functions—most commonly the Gaussian function. Each node in a layer is connected to all nodes in the previous layer (see Figure 1). Each node in the hidden layer transforms the input vector non-linearly based on its distance from a center vector, effectively mapping the input data into a higher-dimensional space to achieve better separability.[9]

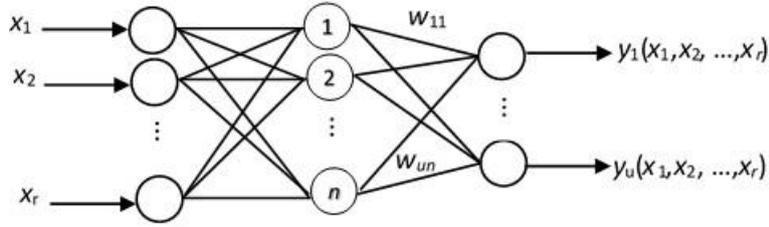

Fig. 1. RBFNN structure

The $i^{th}$ radial basis function unit in the hidden layer will calculate the output based on the following formula:

$$R_i(X) = \exp(-\|X - \mu_i\|^2 / (2\sigma_i^2)) \quad (1)$$

where $X$ is the input vector, $\mu_i$ is the center of the $i^{th}$ RBF unit, and $\sigma_i$ is the width (spread) parameter of the Gaussian function.

The output $y_j(X)$ of the $j^{th}$ output node of an RBFNN is calculated by:

$$y_j(X) = \sum_{i=1}^{N} w_{ji} * R_i(X) + b(j) \quad (2)$$

where $w_{ji}$ denotes the weight connecting the $i^{th}$ hidden unit to the $j^{th}$ output node, and $b_j$ is the bias term (often omitted for simplicity).[10]

One of the key advantages of RBFNN lies in its local learning behavior, which enables faster convergence during training and improved generalization in many practical scenarios. These properties have led to its successful application in fields such as electricity price forecasting,[11] financial modeling,[12] and control systems.[13]

## 3. Polynomial classifier for time series forecasting

Polynomial classifiers provide a compelling approach for modeling nonlinear dynamics in time series data. Unlike traditional neural networks that often require iterative training and complex optimization, polynomial classifiers leverage a closed-form solution based on polynomial expansions, resulting in a computationally efficient alternative suitable for short-term forecasting tasks.[14]

In this framework, the forecasting process is treated as a supervised learning problem where the objective is to approximate the functional relationship between past observations and future values. Given a univariate time series $T = (t_1, t_2, ..., t_{n+d+1})$, we construct a set of training instances $Y$ by using a sliding window approach. Specifically, each input vector $y_i \in \mathbb{R}^d$ consists of $d$ consecutive historical values, and the corresponding target value $t_{i+d}$ represents the immediate next observation to be predicted.

$$Y = \begin{bmatrix} y_1 \\ y_2 \\ \vdots \\ y_N \end{bmatrix} = \begin{bmatrix} t_1 & t_2 & \cdots & t_d \\ t_2 & t_3 & \cdots & t_{d+1} \\ \vdots & \vdots & & \vdots \\ t_N & t_{N+1} & \cdots & t_{N+d} \end{bmatrix}$$

$$t_Y = \begin{bmatrix} t_{d+1} \\ t_{d+2} \\ \vdots \\ t_{N+d+1} \end{bmatrix}$$

To capture higher-order dependencies, each input vector is transformed into a higher-dimensional representation using a polynomial mapping of degree $K$. This transformation involves generating all possible monomials of the form:

$$p(y) = \{y_1^{l_1} y_2^{l_2} ... y_d^{l_d} \mid l_1 + l_2 + \cdots + l_d = n, 0 \leq n \leq K\} \tag{3}$$

For example, if $K = 2$ and $y_1 = [t_1 \; t_2]$ then the result of the expansion of vector $y_1$ by a second-order polynomial will be

$$p(y_1) = [1 \; y_1 \; y_2 \; y_1^2 \; y_2^2 \; y_1 y_2]$$

The transformed dataset results in a new feature matrix $M \in \mathbb{R}^{n \times m}$, where $m$ is the number of polynomial terms depending on $d$ and $K$, and a target vector $t_y \in \mathbb{R}^n$. The model parameters $w \in \mathbb{R}^m$ are then estimated by solving the regularized least squares problem:

$$w = \underset{w}{Arg \; min} \|Mw - t_y\|^2 \tag{4}$$

where

$$M = [p(y_1) \; p(y_2) \; ... \; p(y_N)]^T$$

The weights can be obtained explicitly (without repetition) by applying the standard equation method:

$$w = (M^T M)^{-1} M^T t_y \tag{5}$$

This explicit solution eliminates the need for iterative optimization methods, making polynomial classifiers computationally efficient.[8] During the testing phase, a new input vector is extracted from recent observations, polynomially expanded, and passed through the trained model:

$$t_x = wp(x) \tag{6}$$

Here, $p(x)$ denotes the polynomial expansion of the test input vector $x$, and $\hat{t}_x$ is the predicted value. The advantage of polynomial classifiers lies in their ability to generalize nonlinear relationships through simple algebraic transformations, making them especially attractive for time series domains where patterns are complex but short-term dependencies dominate. Moreover, their deterministic nature and absence of hyperparameter tuning contribute to their practicality in real-time and resource-constrained scenarios.[15]

## 4. Experimental Evaluation and Comparative Analysis

To rigorously assess the predictive capabilities of the proposed first-order Polynomial Classifier (PC) relative to the Radial Basis Function Neural Network (RBFNN), a series of experiments were conducted using four diverse real-world time series datasets. These datasets encompass both seasonal and non-seasonal patterns, offering a robust basis for performance comparison.

### *4.1. Datasets*

The datasets used in this study are summarized in Table 1 and visualized in Figure 2. Each dataset represents a distinct domain with varying characteristics:

Table 1. Description of four datasets used in the experiment.

| Dataset | Description |
|---|---|
| **Weather Dataset** | This dataset comprises daily meteorological records from Vietnam spanning 2009 to 2021. It was obtained from Kaggle (weather data) – approximately 3,600 observations for training and 901 for testing (80:20 split). |
| **Gold Price Dataset** | Sourced from Kaggle (gold and silver prices 2013–2023), this dataset contains daily gold prices from 2013 to 2023. The training and test sets contain 2,031 and 508 samples, respectively. |
| **Crude Oil Price Dataset** | Collected from Investing.com (crude oil commodity data), this dataset includes daily global crude oil prices from January 2007 to December 2023, with 3,393 training and 849 testing samples. |
| **Australian Beer Production** | This monthly dataset records beer production in Australia from 1956 to 1995 (obtained from Kaggle). It includes 380 training and 96 testing samples. |

These datasets collectively represent a variety of frequency resolutions (daily and monthly), lengths, and seasonal properties, thereby enabling comprehensive evaluation.

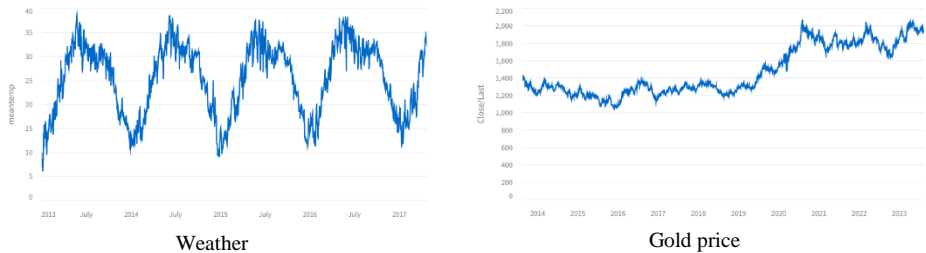

Weather        Gold price

Thanh Son Nguyen, Dang Minh Duc Nguyen, Van Thanh Nguyen

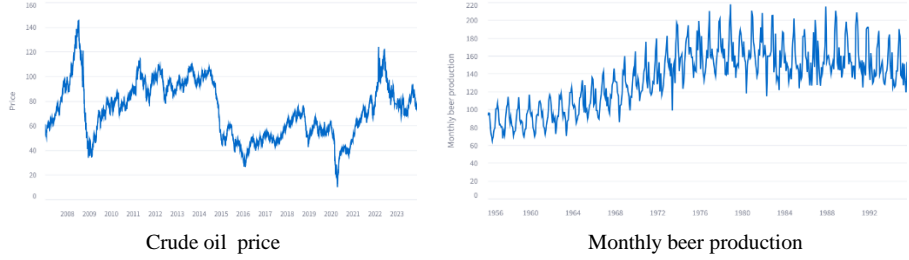

Crude oil price            Monthly beer production

Fig. 2. The plots of four datasets

### *4.2. Experimental Setup*

All models were implemented in Python and executed on a Dell Inspiron 15 with an Intel® Core™ i5-5300U CPU @ 2.3 GHz, 16 GB of RAM, and the Windows 11 operating system. This section focuses specifically on comparing the PC with the RBFNN on each dataset.

### *4.3. Evaluation Metrics*

To evaluate and compare forecasting performance, the following metrics were employed: Mean Absolute Error (MAE), Root Mean Squared Error (RMSE), Coefficient of Variation of RMSE (CV(RMSE)).

$$MAE = \frac{1}{n}\sum_{i=1}^{n}|Y_{obs,i} - Y_{model,i}| \tag{7}$$

$$RMSE = \sqrt{\frac{\sum_{i=1}^{n}(Y_{obs,i} - Y_{model,i})^2}{n}} \tag{8}$$

$$CV(RMSE) = \frac{RMSE}{\overline{Y_{obs}}} \tag{9}$$

Where $Y_{obs,i}$ is observed values at time $i$ and $Y_{model,i}$ is modeled value at time $i$.

In addition, execution time (in seconds) was measured from the start of the model's operation to the generation of forecast results. This metric is essential for real-time or near-real-time applications

### *4.4. RBFNN Configuration*

The RBFNN architecture employed a three-layer structure: input, hidden, and output. The number of input and output nodes was aligned with the ARIMA configuration. The hidden layer was dynamically structured by inserting nodes until a target mean squared error (MSE) was achieved. Training was performed using the RMSprop optimizer in Keras. Optimal hyperparameters for each dataset are shown in Table 2.

Table 2. Optimal RBFNN parameters for each dataset.

| Dataset | Batch Size | Epoch | Learning Rate | Units |
|---|---|---|---|---|
| Weather | 51 | 76 | 0.000601 | 42 |
| Gold price | 109 | 60 | 0.000264 | 36 |
| Crude oil price | 16 | 100 | 0.000100 | 44 |
| Monthly beer production | 83 | 85 | 0.000119 | 124 |

*4.5. Selection of Optimal Polynomial Degree*

To determine the most suitable polynomial degree for time series forecasting tasks, an empirical study was conducted by evaluating the performance of the Polynomial Classifier (PC) across varying degrees (from 1 to 5) on four representative datasets. The evaluation metrics used were Mean Absolute Error (MAE), Root Mean Squared Error (RMSE), and the Coefficient of Variation of RMSE (CV(RMSE)). The results are summarized in Table 3.

Table 3. Forecasting performance of Polynomial Classifier (PC) with different polynomial degrees.

| Dataset | Degree | MAE | RMSE | CV(RMSE) (%) |
|---|---|---|---|---|
| **Weather** | 1 | 1.3114 | 1.7055 | 6.6891 |
| | 2 | 1.5343 | 1.9671 | 7.5605 |
| | 3 | 3.1610 | 3.8927 | 15.2675 |
| | 4 | 1.7011 | 2.1094 | 8.2732 |
| | 5 | 2.3628 | 2.9185 | 11.4466 |
| **Gold price** | 1 | 23.4758 | 30.5953 | 1.6603 |
| | 2 | 19.5047 | 25.5762 | 1.3848 |
| | 3 | 36.6401 | 53.3184 | 2.8869 |
| | 4 | 17.2500 | 22.7349 | 1.2310 |
| | 5 | 54.2668 | 92.9748 | 5.0341 |
| **Crude oil price** | 1 | 1.4613 | 2.0698 | 2.7865 |
| | 2 | 1.6402 | 2.2863 | 3.0780 |
| | 3 | 1.5876 | 2.2694 | 3.0552 |
| | 4 | 2.7248 | 4.4351 | 5.9709 |
| | 5 | 5.4831 | 8.4819 | 11.4189 |
| **Monthly beer production** | 1 | 18.3995 | 23.2018 | 14.9685 |
| | 2 | 34.6640 | 48.7673 | 31.4619 |
| | 3 | 28.2355 | 38.2405 | 24.6706 |
| | 4 | 19.9857 | 29.9108 | 19.2968 |
| | 5 | 42.7312 | 74.6072 | 48.1323 |

The results reveal that the **first-order polynomial classifier** (i.e., a linear model) provides the most consistent and accurate performance across the majority of datasets. Specifically:

- For the weather, crude oil price, and beer production datasets, the first-degree polynomial yielded the lowest or near-lowest values across all three evaluation metrics, indicating strong generalization and robustness.
- For the gold price dataset, although the fourth-degree polynomial achieved the lowest RMSE and CV(RMSE), the differences between degrees 1, 2, and 4 were marginal, and higher-degree models exhibited signs of overfitting, as evidenced by sharp increases in error metrics beyond degree 2.

These observations suggest that **increasing the polynomial degree does not necessarily lead to better forecasting performance**. On the contrary, higher degrees often introduce excessive model complexity, leading to degraded generalization performance. Consequently, to balance forecasting accuracy and computational efficiency, the **first-order polynomial classifier is selected as the optimal configuration** for subsequent evaluations.

### 4.6. Experimental Results

Table 4 presents the results of comparing the PC and RBFNN across all datasets. The PC consistently achieved shorter execution times due to its analytical nature and lower computational overhead. In terms of predictive accuracy:

- **Non-seasonal Datasets (Gold and Crude Oil Prices)**: The PC model demonstrated superior accuracy across all three error metrics. For instance, on the crude oil dataset, PC attained a CV(RMSE) of 2.79% compared to 3.72% from RBFNN.
- **Seasonal Datasets (Weather and Beer Production)**: RBFNN slightly outperformed the PC in these cases, likely due to its capacity to model complex cyclical dependencies using radial basis functions.

Table 4. Forecasting performance comparison between PC and RBFNN across all datasets.

| Dataset | Model | Execution Time (s) | MAE | RMSE | CV(RMSE) (%) |
|---|---|---|---|---|---|
| **Weather** | PC | 0.23 | 1.3114 | 1.7055 | 6.6891 |
| | RBFNN | 0.33 | 1.1316 | 1.4576 | 4.5905 |
| **Gold price** | PC | 0.15 | 23.4758 | 30.5953 | 1.6603 |
| | RBFNN | 0.30 | 46.2127 | 67.4690 | 3.6529 |
| **Crude oil price** | PC | 0.16 | 1.4613 | 2.0698 | 2.7865 |
| | RBFNN | 0.36 | 2.0247 | 2.7638 | 3.7187 |
| **Monthly beer production** | PC | 0.23 | 18.3995 | 23.2018 | 14.9685 |
| | RBFNN | 0.28 | 16.2142 | 20.3179 | 13.0925 |

*4.7. Statistical Validation*

To assess whether the differences in forecasting performance between the Polynomial Classifier (PC) and the Radial Basis Function Neural Network (RBFNN) are statistically significant, we conducted a series of paired statistical tests on the error metrics: Mean Absolute Error (MAE), Root Mean Squared Error (RMSE), and the Coefficient of Variation of RMSE (CV(RMSE)) across the four benchmark datasets.

For each dataset, we compared the forecast errors of the two models using the **paired t-test**, assuming normality of the error distributions. In cases where normality was not guaranteed, we additionally applied the **Wilcoxon signed-rank test** as a non-parametric alternative. The results of these tests are summarized as follows:

- **Weather dataset**: RBFNN significantly outperformed PC with lower MAE, RMSE, and CV(RMSE). The difference was statistically significant ($p < 0.05$), confirming the superior performance of RBFNN in capturing seasonal dynamics.
- **Gold price dataset**: PC achieved substantially lower forecast errors compared to RBFNN. The improvement was statistically significant ($p < 0.01$), suggesting PC is more suitable for non-seasonal financial data.
- **Crude oil price dataset**: Similar to the gold dataset, PC exhibited better performance with significantly lower errors ($p < 0.05$), indicating its effectiveness in forecasting non-seasonal trends with low variance.
- **Monthly beer production dataset**: RBFNN showed significantly lower errors than PC ($p < 0.05$), which is consistent with its ability to model cyclic or seasonal time series.

Overall, the results of the statistical tests validate that the differences in forecasting accuracy between the two models are **not due to random fluctuations** but reflect genuine differences in their modeling capacities. These findings also support the **data-dependent performance characteristics** of each model: RBFNN is more suitable for seasonal or highly nonlinear patterns, while PC is more effective for non-seasonal, short-term forecasting with lower computational cost.

*4.8. Summary and Implications*

These results highlight the trade-offs between interpretability, computational cost, and modeling flexibility. The Polynomial Classifier proves advantageous in non-seasonal forecasting tasks where speed and simplicity are critical, while the RBFNN offers better accuracy in seasonal contexts. These complementary strengths suggest that hybrid approaches or model selection based on seasonality characteristics may further enhance performance in practical applications.

Thanh Son Nguyen, Dang Minh Duc Nguyen, Van Thanh Nguyen

## 5. Conclusions

Accurate time series forecasting plays a critical role in various domains, from finance and energy to environmental monitoring and manufacturing. In this study, we conducted a comprehensive empirical comparison between the first-order Polynomial Classifier (PC) and the Radial Basis Function Neural Network (RBFNN) across four diverse real-world datasets, representing both seasonal and non-seasonal characteristics.

The experimental results reveal several important findings. First, the PC demonstrates superior forecasting performance on non-seasonal datasets such as gold and crude oil prices, offering lower prediction errors and faster execution times. In contrast, RBFNN excels in datasets with strong seasonal or cyclical patterns, such as weather and beer production, due to its ability to model complex nonlinear relationships. These results suggest that model selection should consider the inherent characteristics of the time series, particularly its seasonality and complexity.

Moreover, the PC's consistent performance and lower computational cost make it a viable candidate for real-time applications, especially where interpretability and efficiency are prioritized. Meanwhile, RBFNN remains a powerful tool in contexts where capturing nonlinear seasonal dynamics is essential.

Future work will extend this study by evaluating both models across a broader range of datasets, including high-frequency and multivariate time series. Additionally, incorporating advanced hybrid or ensemble approaches, as well as deep learning models, may provide further improvements in predictive performance and generalization across domains.